\documentclass[runningheads]{llncs}
\usepackage{graphicx}
\usepackage{comment}
\usepackage{amsmath,amssymb} 
\usepackage{color}
\usepackage{bm}
\usepackage{multirow}

\newcommand*\samethanks[1][\value{footnote}]{\footnotemark[#1]}

\usepackage{graphicx}
\usepackage{subfigure}
\usepackage{algorithm}
\usepackage{algorithmic}
\usepackage{caption}

\DeclareMathOperator*{\argmin}{arg\,min}

\begin{document}
\pagestyle{headings}
\mainmatter
\def\ECCVSubNumber{4107}  

\title{Human Correspondence Consensus \\for 3D Object Semantic Understanding} 

\titlerunning{Human Correspondence Consensus}
%
\author{Yujing Lou\thanks{These authors contributed equally.} \and
Yang You\samethanks \and
Chengkun Li\samethanks \and
Zhoujun Cheng \and
Liangwei Li \and
Lizhuang Ma \and
Weiming Wang \and
Cewu Lu\thanks{Cewu Lu is the corresponding author, who is also the member of Qing Yuan Research Institute and MoE Key Lab of Artificial Intelligence, AI Institute, Shanghai Jiao Tong University, China.}
}
\authorrunning{Y. Lou et al.}
%
\institute{Shanghai Jiao Tong University, Shanghai, China \\
\email{\{louyujing,qq456cvb,sjtulck,blankcheng\}@sjtu.edu.cn\\
\{liliangwei,ma-lz,wangweiming,lucewu\}@sjtu.edu.cn}}

\maketitle

\begin{abstract}
  Semantic understanding of 3D objects is crucial in many applications such as object manipulation. However, it is hard to give a universal definition of point-level semantics that everyone would agree on. We observe that people have a consensus on semantic correspondences between two areas from different objects, but are less certain about the exact semantic meaning of each area. Therefore, we argue that by providing human labeled correspondences between different objects from the same category instead of explicit semantic labels, one can recover rich semantic information of an object. In this paper, we introduce a new dataset named \textbf{C}orres\textbf{P}ondence\textbf{Net}. Based on this dataset, we are able to learn dense semantic embeddings with a novel geodesic consistency loss. Accordingly, several state-of-the-art networks are evaluated on this correspondence benchmark. We further show that \textbf{C}orres\textbf{P}ondence\textbf{Net} could not only boost fine-grained understanding of heterogeneous objects but also cross-object registration and partial object matching.
\end{abstract}

\section{Introduction}


Object understanding~\cite{leng20163d,mo2019partnet,zhou2019semantic} is one of the holy grails in computer vision. Being able to fully understand object semantics is crucial for various applications such as self-driving~\cite{bojarski2016end,paden2016survey} and attribute transfer~\cite{liao2017visual}. Recently, significant advances have been made in both category-level and instance-level understanding of objects~\cite{chang2015shapenet,kundu20183d}. However, these datasets all require explicit semantic labels with an ``oracle'' definition and are not suitable for point-level understanding of objects.

One of the key problems with object semantic understanding lies in the ambiguous definitions of semantics. In the past decades, researchers have proposed keypoints~\cite{leutenegger2011brisk,lin2014microsoft,salti2015learning,suwajanakorn2018discovery,you2020keypointnet} and skeletons~\cite{au2008skeleton} to explicitly define object semantics. These methods have made success in tasks like human body parsing~\cite{kalayeh2018human}, however, it is hard or even impossible to give consistent definitions of keypoints or skeletons for a general object. Recently, part based representations of objects are also adopted by researchers~\cite{chang2015shapenet,yi2016scalable,mo2019partnet,huanglocaldes}, where an object is decomposed into semantic parts by experts, with a predefined semantic label on each part. The above methods all impose an explicit definition of object semantics, which is inevitably biased or flawed since different people may hold different opinions of what the semantics of an object are.  

\begin{figure}[t]
\begin{center}
  \includegraphics[width=0.7\linewidth]{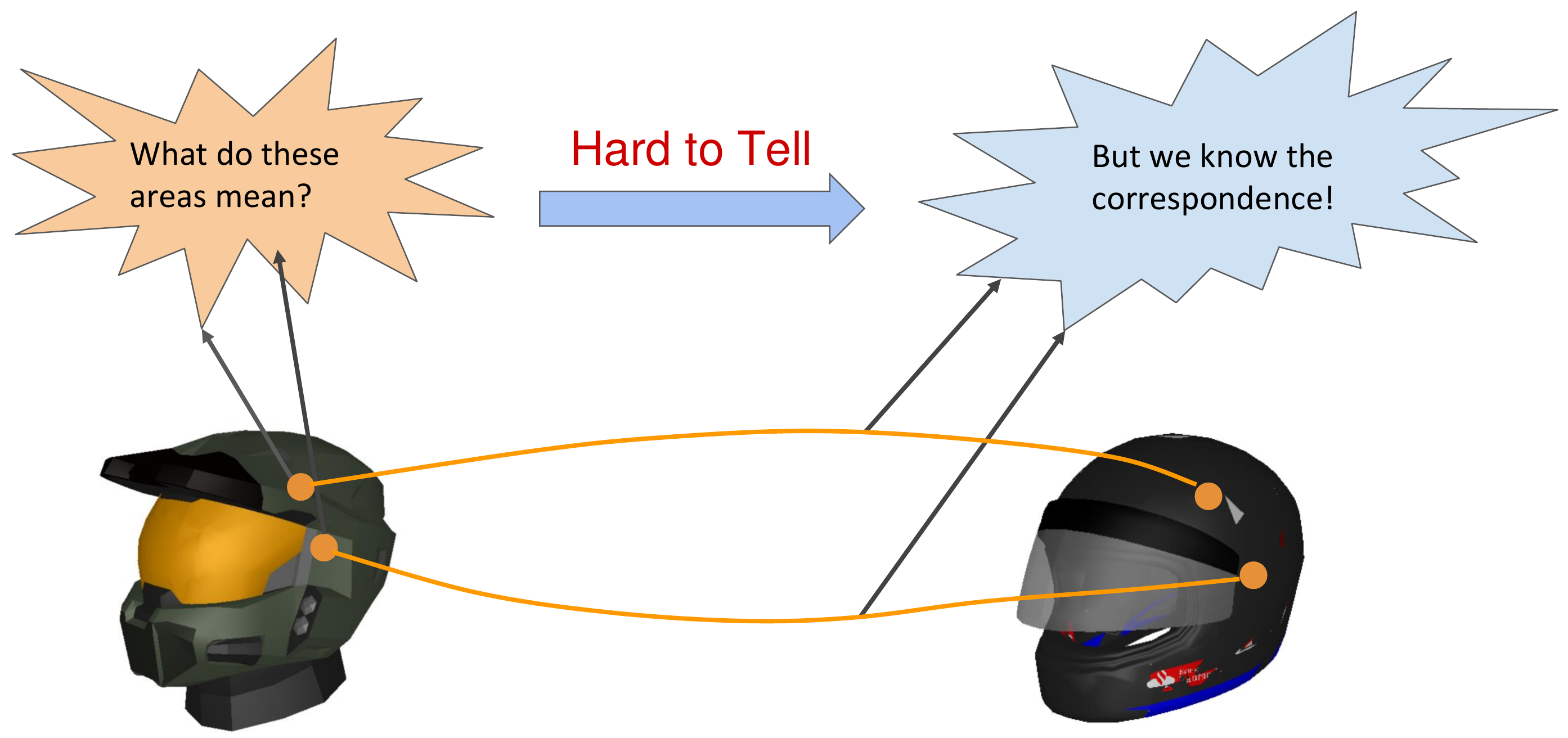}
\end{center}
\caption{We observe that it is hard to tell the exact meanings of some areas on an object, while correspondences between different objects are clear.}
\label{fig:intro}
\end{figure}

In this paper, we explore a brand new way to deal with this vagueness in object semantic understanding. Instead of explicitly giving semantic components and labels, we leverage human semantic correspondence consensus between objects to implicitly infer their semantic meanings. This is based on the observation that while it is hard to tell the exact meanings of some sub-object areas, almost everyone would agree on their semantic correspondence across different objects, as shown in Figure~\ref{fig:intro}. Consequently, comprehensive object understanding can be achieved by collecting multiple unambiguous semantic correspondences from a large population.



\begin{figure*}[t]
    \centering
    \includegraphics[width=\linewidth]{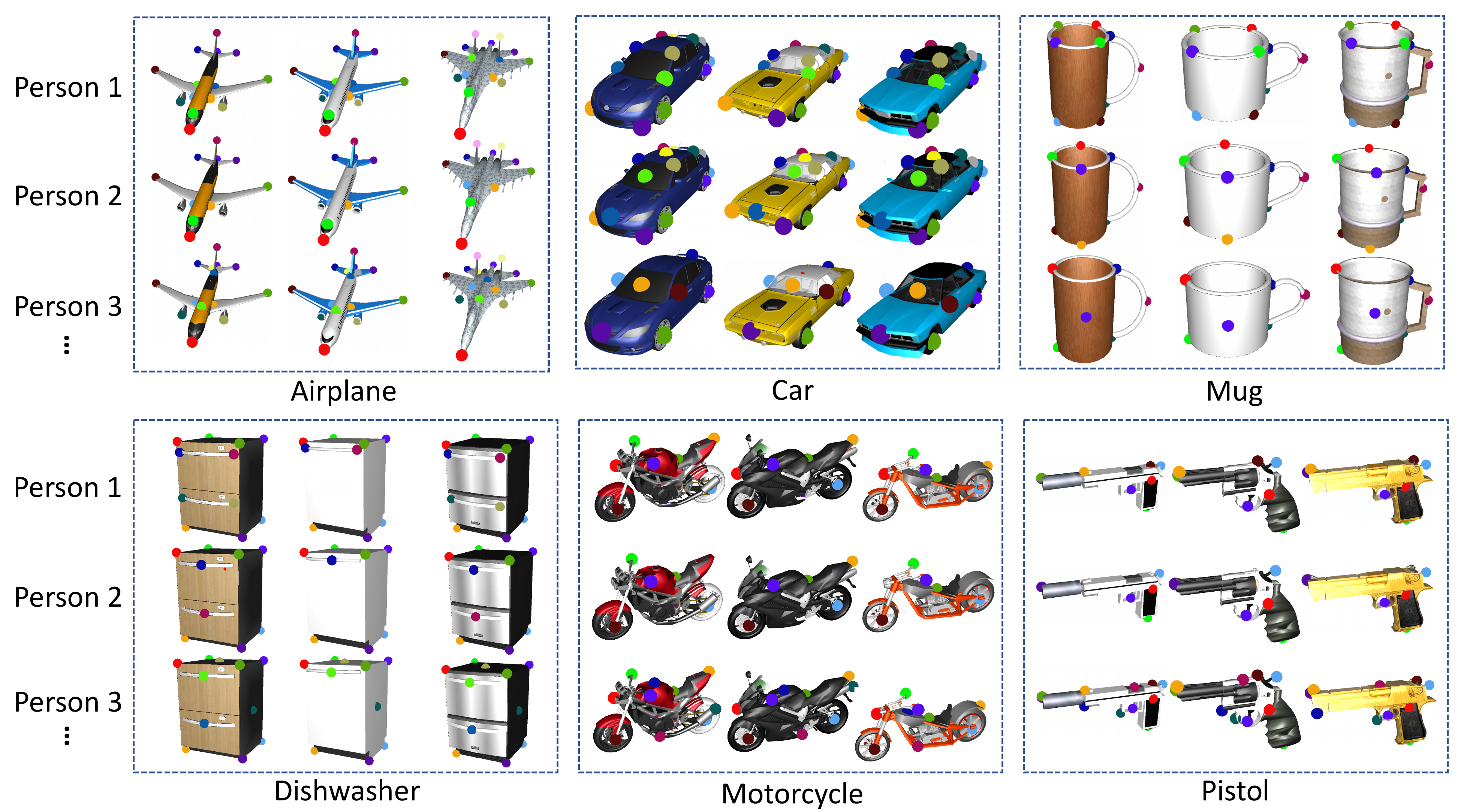}
    \caption{\textbf{CPNet dataset}. Each person annotates multiple sets of corresponding points. Points in the same correspondence set are in the same color. It can be seen that people could have his/her own understanding of semantic points as long as they are consistent across different models within the same category.}
    \label{fig:overall}
\end{figure*}

To that end, we introduce \textbf{C}orres\textbf{P}ondence\textbf{Net}  (CPNet): a \textit{diverse} and \textit{high-quality} dataset on top of ShapeNet~\cite{chang2015shapenet} with \textit{cross-object}, \textit{point-level} 3D semantic correspondence annotations. In this dataset, every annotator gives multiple sets of semantic-consistent points across different intra-class objects, which we call ``correspondence sets'', as shown in Figure~\ref{fig:overall}. 


Using these correspondence sets, we are able to obtain dense semantic embeddings of an object, perform cross-object semantic registration and partial-to-complete object matching.
For dense semantic embedding prediction, a new benchmark with mean Geodesic Error (mGE) is proposed. We leverage a novel geodesic consistency loss to learn this embedding, where points in the same correspondence set are pulled together in the embedding space, while points across different correspondence sets get pushed according to their average geodesic distances. By considering geodesic relationships between different correspondence sets, points with similar semantics are more likely to be grouped together in the embedding space. 

In summary, our key contributions are as follows:
\begin{itemize}
    \item We explore a new way towards 3D object semantic understanding of objects, where explicit definitions are avoided but point-level semantic correspondences across heterogeneous objects are leveraged.
    \item We introduce \textit{CPNet}, the first human correspondence consensus based dataset for 3D object understanding, which contains 100K+ high-quality semantic-consistent points.
    \item Based on \textit{CPNet}, we show several applications include dense semantic embedding prediction, cross-object registration and partial-to-complete object matching. We also propose a new benchmark on dense semantic embedding prediction.
\end{itemize}

The rest of this paper is organized as follows: in Section~\ref{sec:related}, we discuss some related works; in Section~\ref{sec:corr}, we briefly discuss the importance of human correspondence consensus and introduce our dataset with our annotation methods; in Section~\ref{sec:dense}, we discuss a detailed method on learning dense embeddings based on our dataset; in Section~\ref{sec:app}, we show some other applications that are naturally driven by our dataset.

\section{Related Work}
\label{sec:related}
\paragraph{Datasets on Semantic Analysis}
Big data and deep learning have witnessed several large 2D/3D datasets these years aiming to parse semantic information from objects. In the world of 2D images, SPAIR-71k~\cite{min2019spair} proposes a large-scale dataset with rich annotations on viewpoints, keypoints and segmentations, which is mainly used for semantic matching between different images. Recently, Ham et al.~\cite{ham2017proposal} and Taniai et al.~\cite{taniai2016joint} have introduced datasets with groundtruth correspondences. Since then, PF-WILLOW and PF-PASCAL~\cite{ham2017proposal} have been used for evaluation in many works. In addition, plenty of datasets on human pose analysis~\cite{andriluka14cvpr,PoseTrack} have been proposed recently. These 2D image datasets have their advantages in that they are relatively large and pertain diversity across different scenes and objects. 

On the other hand, there exists a rich set of 3D model datasets that try to directly process meshes or point clouds. There are generally two types of them: ones that focus on rigid models and some others that focus on non-rigid models. For rigid model analysis, ShapeNet Core 55~\cite{chang2015shapenet} is proposed to help object-level classification while ShapeNet part dataset~\cite{yi2016scalable} pushes it one step forward with intra-object part classification. As a followup, PartNet~\cite{mo2019partnet} comes up with a much more complete and manually defined hierarchical structures of parts. Alternatively, dataset proposed by Dutagaci et al.~\cite{dutagaci2012evaluation} focuses on sparse semantic keypoints on objects. For non-rigid (deformable) models, FAUST~\cite{bogo2014faust} and TOSCA~\cite{bronstein2008numerical} provide dense correspondence labels for humans and animals, respectively. These methods leverage the clear anatomy structure underlying humans and animals and can be applied to pose transfer, pose synthesis, etc.


\paragraph{Methods on Object Semantic Understanding}
In the last decade, plenty of methods have been proposed to find semantic correspondences between paired images. Earlier methods like Okutomi et al.~\cite{okutomi1993multiple}, Horn et al.~\cite{horn1993determining} and Matas et al.~\cite{matas2004robust} propose to find semantic correspondences within the same scene. Semantic flows like SIFT flow~\cite{liu2010sift} and ProposalFlow~\cite{ham2017proposal} further explore to find dense correspondence across different scenes. Kulkarni et al.~\cite{kulkarni2019canonical} and Zhou et al.~\cite{zhou2016learning} utilize a synthesis 3D model as a medium to enforce semantic cycle-consistency. Florence et al.~\cite{florence2018dense} and Schmidt et al.~\cite{schmidt2016self} leverage an unsupervised method to learn consistent dense embeddings across different objects.

When it comes to the domain of 3D shapes, Blanz et al.~\cite{blanz1999morphable} and Allen et al.~\cite{allen2003space} are the pioneers on finding 3D correspondence between human faces and bodies. Recently, 3D dense semantic correspondence has been boosted by a variety of deep learning methods. Halimi et al.~\cite{halimi2018self}, Groueix et al.~\cite{groueix20183d} and Roufosse et al.~\cite{roufosse2019unsupervised} propose unsupervised methods on learning dense correspondences between humans and animals. Deep functional dictionaries~\cite{sung2018deep} gives a small dictionary of basis functions for each shape, a dictionary whose span includes the semantic functions provided for that shape. Perhaps, closest to this paper, is the method of Huang et al.~\cite{huanglocaldes}. It utilized expert-defined corresponding shape parts to generate a synthetic dense point correspondence dataset and then extracts local descriptors by a neural network. However, it is ambiguous to clearly define object parts while we do not leverage any expert-defined part labels. In addition, their assumption of dense one-to-one correspondence within the same part
fails in many common objects.

\section{CorresPondenceNet}
\label{sec:corr}

Understanding semantics from arbitrary objects is of great importance. However, explicitly expressing semantics in a well defined format is extremely hard as the definition of semantics is vague and diverse. 

We observe that people are pretty sure about the correspondence between two areas but less sure about what each area means in semantics. As shown in Figure~\ref{fig:intro}, almost everyone would agree on the lined correspondences between two helmets. However, it is hard to tell the exact semantic meanings of the colored areas.

Unlike all previous methods where an explicit definition of keypoints or parts is given, we instead focus on sparse correspondences annotated by humans, based on the assumption that all the corresponding points labeled by the same person share the same semantic meaning. 

Therefore, we propose a new dataset called \textbf{C}orres\textbf{P}ondenceNet (CPNet). CPNet has a collection of 25 categories, 2,334 models based on ShapeNetCore with 104,861 points. Each model is annotated with a number of semantic points from multiple annotators, as shown in Figure~\ref{fig:overall}. Unlike other 2D or 3D keypoint datasets which manually set a keypoint template and let annotators to follow, semantic points in our dataset are not deliberately defined by anyone. The key is that every annotator can have his/her own understanding of semantic points, as long as they are consistent across different models within the same category. In the following subsections, we discuss how we collect models, how we annotate models and annotation types in details. Table~\ref{tab:statistics} gives the detailed statistics of our dataset.

\subsection{Dataset Collections}
Our dataset is based on ShapeNetCore~\cite{chang2015shapenet}. ShapeNetCore is a subset of the full ShapeNet dataset with single clean 3D models and manually verified category and alignment annotations. There are 51,300 unique 3D models from 55 common object categories in ShapeNetCore. We select 25 categories that are mostly seen in daily life to build our dataset. To keep a balanced dataset, for each category we keep at most 100 models. For those categories with less than 100 models, all the models are selected.

\begin{table}[!t]

\begin{center}
\resizebox{\textwidth}{!}{
\begin{tabular}{l|c c c c c c c c c c c c c}

\hline
~  & Airplane & Bathtub & Bed & Bench & Bottle & Bus & Cap & Car & Chair & Dishwasher & Display & Earphone & Faucet \\
\hline
N\textsubscript{P} & 5527 & 6033 & 6464 & 5421 & 4489 & 6404 & 949 & 7938 & 6140 & 5343 & 4509 & 904 & 1612 \\
N\textsubscript{A} & 10 & 10 & 10 & 10 & 10 & 10 & 10 & 10 & 10 & 10 & 10 & 10 & 10 \\
N\textsubscript{M} & 100 & 100 & 100 & 100 & 100 & 100 & 38 & 100 & 100 & 77 & 100 & 58 & 100 \\
\hline
C\textsubscript{min} & 35 & 40 & 40 & 30 & 41 & 50 & 20 & 64 & 50 & 60 & 20 & 14 & 10 \\
C\textsubscript{med} & 54 & 60 & 60 & 50 & 45 & 64 & 25 & 80 & 70 & 70 & 50 & 15 & 15 \\
C\textsubscript{max} & 72 & 96 & 80 & 70 & 46 & 81 & 30 & 82 & 78 & 84 & 51 & 21 & 22 \\
\hline
\end{tabular}}
\end{center}   
\begin{center}
\resizebox{\textwidth}{!}{
\begin{tabular}{l|c c c c c c c c c c c c|c}
\hline
~  & Guitar & Helmet & Knife & Lamp & Laptop & Motorcycle & Mug & Pistol & Rocket & Skateboard & Table & Vessel & All\\
\hline
N\textsubscript{P} & 2832 & 1500 & 2109 & 1683 & 2987 & 3878 & 7668 & 3358 & 2315 & 3822 & 4008 & 5214 & 104861 \\
N\textsubscript{A} & 10 & 10 & 10 & 10 & 10 & 10 & 10 & 10 & 10 & 10 & 10 & 10 & - \\
N\textsubscript{M} & 100 & 95 & 100 & 100 & 100 & 100 & 100 & 100 & 66 & 100 & 100 & 100 & 2334 \\
\hline
C\textsubscript{min} & 19 & 27 & 10 & 13 & 20 & 30 & 66 & 17 & 21 & 20 & 39 & 40 & - \\
C\textsubscript{med} & 30 & 35 & 12 & 15 & 30 & 40 & 77 & 35 & 32 & 40 & 40 & 54 & - \\
C\textsubscript{max} & 32 & 37 & 15 & 21 & 36 & 40 & 78 & 41 & 49 & 43 & 44 & 56 & - \\
\hline
\end{tabular}}
\end{center} 
\caption{\textbf{CPNet statistics.} N\textsubscript{P} gives the number of annotated points of each category; N\textsubscript{A} gives the number of annotators for each category; N\textsubscript{M} is the number of models in each category; C\textsubscript{min}, C\textsubscript{med}, C\textsubscript{max} give minimum, median and maximum number of correspondence sets per instance in each category.}
\label{tab:statistics}
\end{table}

\subsection{Annotation Process}
We hire 80 professional annotators in total. Each model is annotated by at least 10 persons to enrich the dataset.

\begin{figure}[t]
\begin{center}
\subfigure[]{
\begin{minipage}{.47\textwidth}
  \begin{minipage}{.5\textwidth}
  \centering
  \includegraphics[width=.7\linewidth]{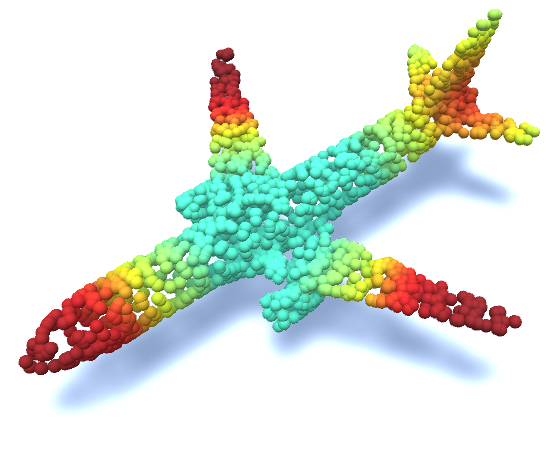}
  \end{minipage}%
  \begin{minipage}{.5\textwidth}
  \centering
  \includegraphics[width=.7\linewidth]{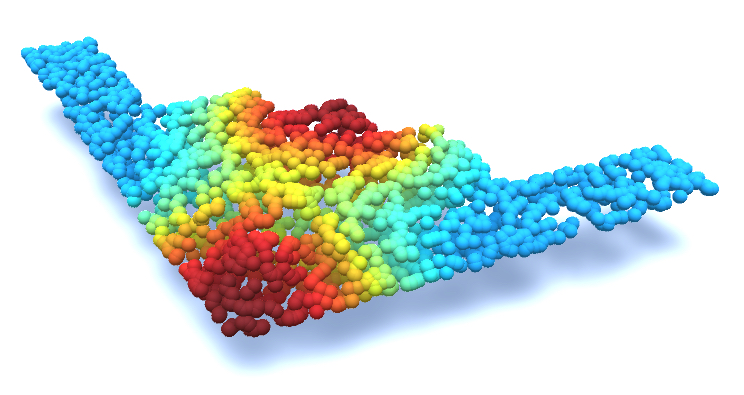}
  \end{minipage}
  \label{fig:coverage}
\end{minipage}
}
\subfigure[]{
\begin{minipage}{.47\textwidth}
  \centering
  \includegraphics[width=\linewidth]{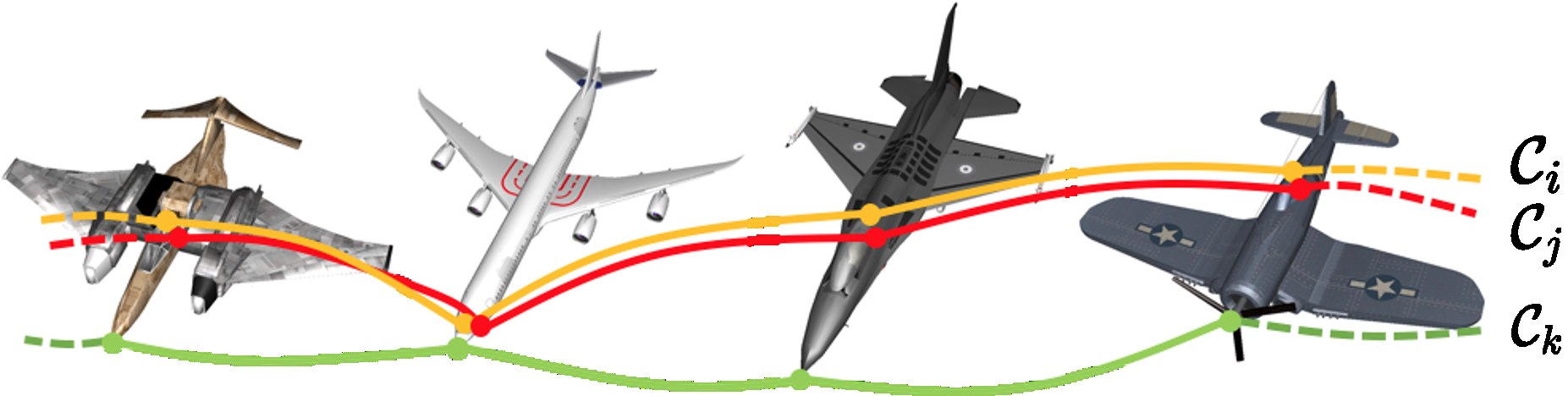}
  \label{fig:corr}
\end{minipage}
}
\end{center}
\caption{(a) Example coverage heatmaps. Red regions indicate frequent annotations while blue means the opposite. We encourage annotators to annotate on blue regions. (b) \textbf{Correspondence sets across different airplanes.} $\mathcal{C}_i$, $\mathcal{C}_j$ and $\mathcal{C}_k$ denote three semantic correspondence sets respectively.}
\end{figure}

\paragraph{Template Creation}
 For each category, every annotator is allowed to create $1$ to $6$ templates with his/her own understanding of semantic points. To ensure a broader range of point coverage, we plot a heatmap for each template to indicate which region has been marked often by others. Annotators are encouraged to mark semantic points in those regions that are less explored. As shown in Figure~\ref{fig:coverage}, red regions indicate frequent annotations while blue means the opposite. Therefore, annotators should avoid red regions in order to get a better coverage.
 
 Templates are then listed to guide the annotations of rest models, so that he/she is able to keep the consistency. Consider an airplane as an example, if one annotator marks the nose as No.1 semantic point, then he/she is supposed to mark all the noses on other airplanes as No.1. It does not matter if another annotator marks the nose as No.2 semantic point, or even neglecting it, as long as the annotator keeps his/her own rules across all the models. For a certain point that does not exist on all the models such as a point of propeller, one can just skip the models without it. The annotator is free to choose any points from his/her perspective.

Each annotator is asked to mark at most 16 semantic points per model. All points are annotated on raw meshes, which is more accurate than those annotated on point clouds. Moreover, it is straightforward to extend these annotations to point clouds by sampling from the mesh while fixing the locations of semantic points.

\paragraph{Handling Symmetries}
In  case  of  any  central/rotational symmetry,  we  extend  our  single  semantic  point $p_{i,j}^{(n)}$  to  a  single  \textit{hyperpoint}, which contains all the points that are centrally/rotationally symmetric. This step is done manually by marking out those symmetric points. During training, \textit{hyperpoint} are treated as normal points.  When generating a positive/negative point pair, we randomly sample a point within the \textit{hyperpoint}.

\paragraph{Cross Validation}
As we mentioned before, we do not define semantic labels. However, this makes strict vetting process impossible. In order to make our dataset trustworthy, we introduce a cross-validation process. To be more specific, for each annotated correspondence, we ask at-least ten other annotators to verify if it is reasonable or not. If more than 80\% annotators agree it is reasonable, then this correspondence is kept, otherwise it is rejected. The rationale for cross-validation lies in our prior that most people have a consistent common sense on whether a given semantic correspondence exists across different objects.

\subsection{Annotation Type} 
\label{sec:annot}
Denote all the models as $\mathbf{M}=\{\mathcal{M}_i\}$, where ${\mathcal{M}_i}$ represents a single model. Each mdoel $\mathcal{M}_i$ is associated with a set of semantic points $\mathcal{P}_i = \{p^{(n)}_{i,j}\}$ where $i, j, n$ denote the $j$-th semantic point of the $n$-th annotator on the $i$-th model.

In addition, we ask each annotator to give consistent points across different models, so that $p^{(n)}_{i_1,j}$ and $p^{(n)}_{i_2,j}$ have the same semantic meaning. Therefore, we define a set of correspondence sets $\Omega = \{\mathcal{C}_j|j=1,\cdots,N_\Omega\}$, where each correspondence set $\mathcal{C}_j = \{p_{i,j}|i=1,\cdots,N_\mathbf{M}\}$ contains all the points with the same semantic label. Note that we dropped the index of the annotator since distinct point correspondence from the same person can be treated the same as those from different persons.

Each annotated point contains attributes about (1) $xyz$ coordinate, (2) color, (3) face index and (4) $uv$ coordinate. By providing these attributes, methods based on either point clouds or meshes can be applied easily.

We thus release four different versions of our correspondence dataset for those who are interested: 1) correspondences without any symmetries; 2) correspondences with only central symmetries; 3) correspondences with only rotational symmetries; 4) correspondences with both central and rotational symmetries.


\section{Learning Dense Semantic Embeddings}
\label{sec:dense}

We now propose a method on learning dense semantic embeddings from human labeled correspondences across various intra-class models.

\subsection{Problem Statement}

Given a set of 3D models $\mathbf{M} =\{\mathcal{M}_i| i=1,\cdots,N_{\mathbf{M}}\}$ and a set of correspondence sets $\Omega = \{\mathcal{C}_j|j=1,\cdots,N_\Omega\}$ defined in Section~\ref{sec:annot},
our goal is to produce a set of pointwise embeddings for each model $\mathcal{M}_i$. The embeddings encode semantic information across different models and points with similar semantics are close in embedding space. We define $f$ as an embedding function, such that $f(p)$ gives the embedding for point $p$ on the model. In practice, we approximate $f$ with a deep neural network and explain how to optimize $f$ as follows.

\subsection{Method Details}
\begin{figure}[h]
\begin{center}
  \includegraphics[width=0.8\linewidth]{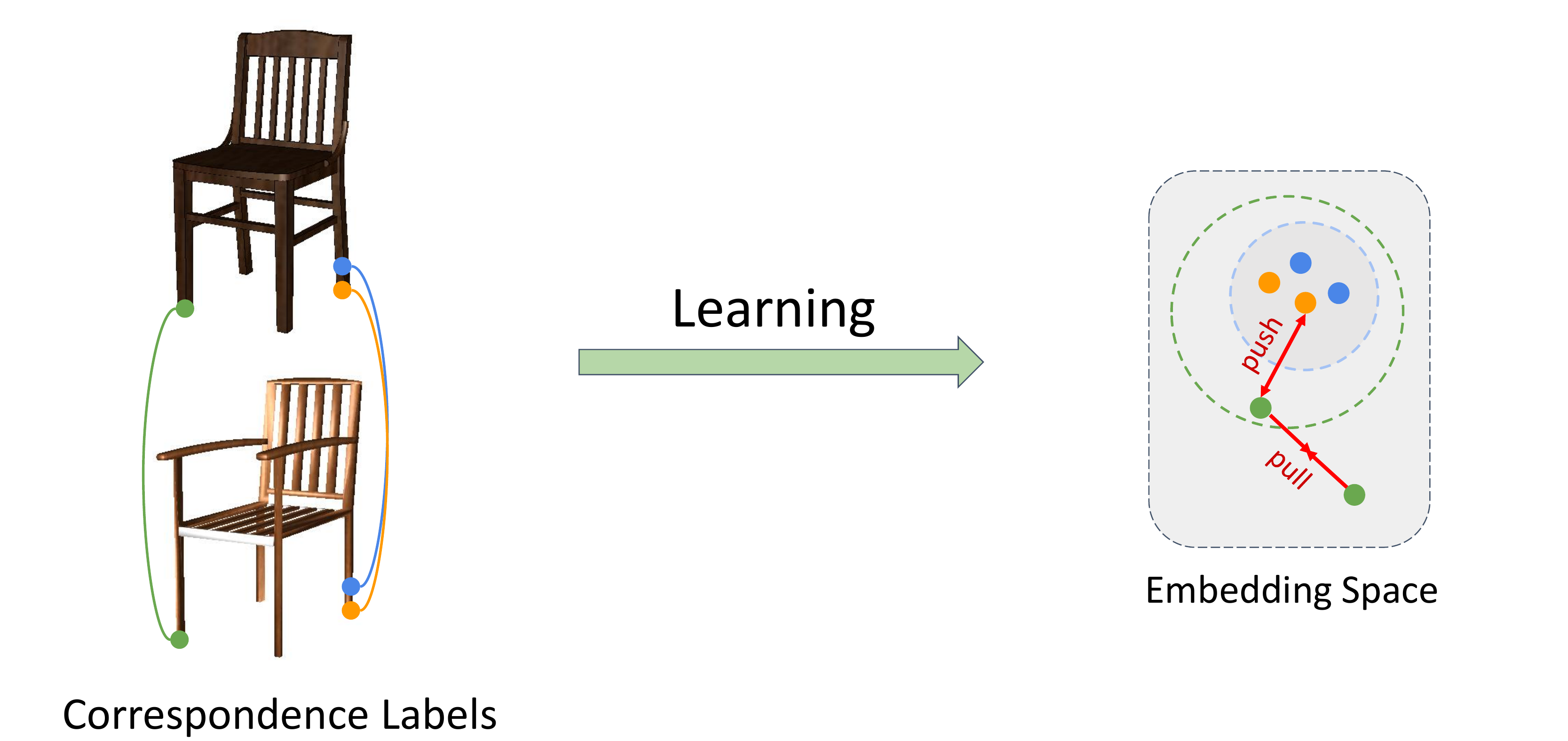}
\end{center}
\caption{Given correspondence sets, we pull the points in the same correspondence set and push points from different correspondence sets adaptively, according to their average geodesic distances. The blue and orange correspondence sets are close so that they can stay close in embedding space, while the orange and green ones are far away in average geodesic distance so their embeddings are pushed further from each other.}
\label{fig:pp}
\end{figure}


 \paragraph{Pull Loss}
 It is natural to come up with a pull loss since we would like to ensure the semantic consistency within every correspondence set. As illustrated in Figure~\ref{fig:corr}, the points with the same color belong to the same correspondence set and reveal similar semantic information.
 For one specific correspondence $\mathcal{C}_k$ like the green line shown in Figure~\ref{fig:corr}, we aim to pull the embedding vectors of the points within it. Any two of points in the same correspondence set form a positive pair. The pairwise embedding distances are then summed over all positive pairs to form our pull loss:
\begin{align}
\begin{split}
\label{eq:pull}
    L_{pull} = \frac{1}{N_{pos}}\sum_{k} \sum_{p, q\in \mathcal{C}_k, p \neq q}\|f(p) - f(q)\|_2,
\end{split}
\end{align}
where $N_{pos}$ is the number of all possible positive point pairs. 

\paragraph{Geodesic Consistency Loss}
The pull loss in Equation~\ref{eq:pull} enforces the points in the same correspondence set to have similar embeddings. However, there is a trivial solution where $f$ outputs a constant embedding (e.g. $\mathbf{0}$) for all points, which is a global optimum when minimizing $L_{pull}$ only. Such a trivial solution is due to the ignorance of an important principle: we ought to ensure that those points with distinct semantics to have a large embedding distance. Therefore, a push loss guided by geodesic consistency is proposed to fulfill this goal. 
We leverage a prior to determine whether two different correspondence sets have distinct semantics: if all pairs of points from these two sets have large geodesic distances on models, they are more likely to reveal different semantic information.

Based on this insight, we design a distance measure $\mathbf{d}$ for a pair of correspondence sets $\mathcal{C}_i$ and $\mathcal{C}_j$:
\begin{align}
\begin{split}
    \mathbf{d}(\mathcal{C}_i, \mathcal{C}_j) = \frac{1}{N_{\mathbf{M}}}\sum_k\sum_{ p,q\in\mathcal{M}_k}\mathbf{d}_{geo}(p, q),\ 
    \textit{s.t. }p \in \mathcal{C}_i, q \in \mathcal{C}_j,
\end{split}
\end{align}
where $\mathbf{d}_{geo}(p, q)$ is the geodesic distance between point $p$ and $q$. This distance measure $\mathbf{d}$ represents the average geodesic distance between point pairs from two correspondence sets.

Then, the push loss can be written as,
\begin{align}
\label{eq:push}
\begin{split}
    L_{push} =  \frac{1}{N_{neg}}\sum_{i\neq j}\sum_{p\in\mathcal{C}_i }\sum_{q\in  \mathcal{C}_j } \max \{0,\mathbf{d}(\mathcal{C}_i, \mathcal{C}_j) - \|f(p) - f(q)\|_2\} ,
\end{split}
\end{align}
 where $N_{neg}$ is the number of all possible negative pairs formed by points from different correspondence sets.

In Equation~\ref{eq:push}, the push loss is only activated when $\|f(p) - f(q)\|_2$ is smaller than $\mathbf{d}(\mathcal{C}_i, \mathcal{C}_j)$. In other words, the larger $\mathbf{d}(\mathcal{C}_i, \mathcal{C}_j)$ is, the further $f(p)$ and $f(q)$ are separated in the embedding space. 
This is based on the observation that some points in two correspondence sets may have similar semantic information (like the red and orange lines in Figure~\ref{fig:corr}) while some have totally different meanings (like the orange and green lines in Figure~\ref{fig:corr}). Therefore, only for those correspondence sets with a large average geodesic distance, a large distance between their embeddings is expected. 




Our final loss is, 
\begin{align}
\label{eq:final}
    L = L_{pull} + \lambda L_{push},
\end{align}
where $\lambda$ is a weight factor. Our method is summarized in Figure~\ref{fig:pp}.

\paragraph{Hard Negative Mining}
In practice, negative pairs to be pushed are combinatorially more than positive pairs to be pulled, since negative pairs are sampled from different correspondence sets. In such case, we borrow the idea from ~\cite{dalal2005histograms} to utilize hard negative mining. Within each batch, only those negative pairs with smallest embedding distances are taken into consideration, matching the number of positive pairs.

\subsection{Mean Geodesic Error}
\begin{algorithm}[h!]
    \caption{mean Geodesic Error calculation} 
    \label{algo:mgd}
    \begin{algorithmic}
        \STATE\textbf{Input}: model set $\Omega$, an embedding function $f$ to be evaluated
        \STATE\textbf{Output}: mean Geodesic Error (mGE) $\varepsilon$ of $f$
        \STATE$\varepsilon = 0$
        \FOR{$\mathcal{C}_i$ \textbf{in} $\Omega$}
            \FOR{$p$, $q$ \textbf{in} $\mathcal{C}_i$}
                    \STATE $x=\argmin_{x\in \mathcal{M}_q}\|f(x) - f(p)\|_2$, where
                    \STATE \quad\quad $\mathcal{M}_q$ denotes the model that point $q$ lies on.
                    \STATE $\varepsilon = \varepsilon + \mathbf{d}_{geo}(q, x)$
            \ENDFOR
        \ENDFOR
        \STATE $\varepsilon = \frac{\varepsilon}{N_{\Omega} N_{\mathbf{M}}^2}$
    \end{algorithmic}
\end{algorithm}

Since we are dealing with a new dense embedding prediction task, existing metrics on classification or part segmentation can not benchmark it well. 
Therefore, we introduce mean Geodesic Error (mGE), a new metric on dense correspondence, to evaluate predicted semantic embeddings. Unlike mean Euclidean Error that is used in Huang at el.~\cite{huanglocaldes}, geodesic distance is more suitable for 3d objects as it is restricted to lie on object surfaces. mGE is calculated individually for each category and measures how well the generated embedding vectors fit with annotated correspondence sets. We also provide results for mean Euclidean Error in the supplementary material.
Algorithm \ref{algo:mgd} presents the calculation procedure of mGE for a given embedding function $f$. Intuitively, for each annotated points on a model, we find their corresponding points that minimize the embedding distance on other models. After that, the geodesic distances between these points and human labeled corresponding points are accumulated. It is easy to verify that if all the embeddings are identical within the same correspondence set but are distinct across different correspondence sets, $\text{mGE} = 0$, which means that the predicted semantic embeddings are consistent with human labels.

\subsection{Experiments}

\begin{figure*}[ht]
\begin{center}
  \includegraphics[width=\linewidth]{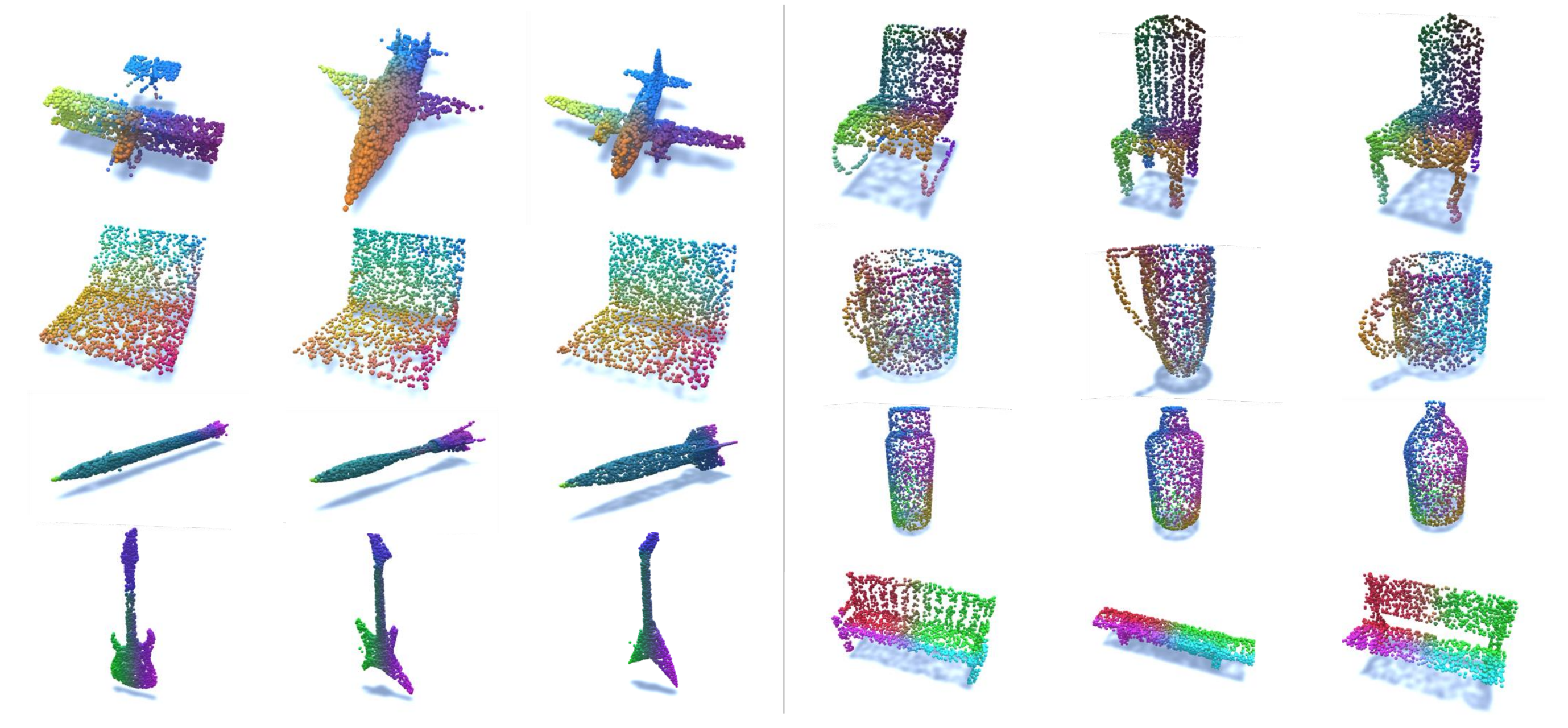}
\end{center}
\caption{\textbf{Predicted semantic embeddings for PointConv.} Same colors indicate similar embeddings.}
\label{fig:emb1}
\end{figure*}

\begin{table}[t]
\begin{center}
\resizebox{\textwidth}{!}{
\begin{tabular}{l|c c c c c c c c c c c c c}
\hline
~  & Airplane & Bathtub & Bed & Bench & Bottle & Bus & Cap & Car & Chair & Dishwasher & Display & Earphone & Faucet \\
\hline
PointNet & 0.063&	0.141&	\textbf{0.078}&	0.066&	0.090&	\textbf{0.055}&	0.093&	0.070&	\textbf{0.088}&	\textbf{0.103}&	\textbf{0.071}&	0.151&	0.163 \\
PointNet++ & 0.053&	0.170&	0.118&	0.071& 0.138	&	0.118&	0.123&	0.075&	0.114&	0.148&	0.112& 0.122	&	0.179 \\
RS-Net & 0.052&	0.153&	0.121&	0.091&	\textbf{0.082}&	0.059&	0.101&	\textbf{0.064}&	0.097&	0.145&	0.081&	0.115&	0.167 \\
PointConv &0.053	&0.133	&0.128&	0.072&	0.100&	0.076&	0.121&	0.079&	0.126&	0.144&	0.085&	0.103&	0.161 \\
\hline
DGCNN & \textbf{0.046}&	\textbf{0.118}&	0.125&	\textbf{0.058}&	0.088&	0.060&	\textbf{0.085}&	0.073&	0.106&	0.116&	0.086&	\textbf{0.091} &	\textbf{0.143}\\
GraphCNN & 0.069&	0.153&	0.126&	0.089&	0.166&	0.099&	0.122&	0.112&	0.147&	0.157&	0.132&	0.136&	0.163 \\
\hline
Minkowski & 0.085&	0.149&	0.150&	0.112&	0.147&	0.113&	0.155&	0.102&	0.162&	0.177&	0.179&	0.116&	0.185\\
\hline
SHOT & 0.230&	0.485&	0.580&	0.568&	0.380&	0.410&	0.340&	0.386&	0.508&	0.515&	0.430&	0.495&	0.258 \\
\hline
Random & 0.308&	0.492&	0.564&	0.544&	0.431&	0.404&	0.484&	0.401&	0.515&	0.507&	0.483&	0.599&	0.355 \\
\hline
\end{tabular}}
\end{center}   
\begin{center}
\resizebox{\textwidth}{!}{
\begin{tabular}{l|c c c c c c c c c c c c |c}
\hline
~  & Guitar & Helmet & Knife & Lamp & Laptop & Motorcycle & Mug & Pistol & Rocket & Skateboard & Table & Vessel & Average\\
\hline
PointNet & 0.066&	0.169&	0.066&	\textbf{0.221}&	0.163&	0.085&	0.072&	\textbf{0.091}&	0.151&	\textbf{0.059}&	\textbf{0.042}&	\textbf{0.101}&	0.101\\
PointNet++ & 0.083&	0.180&	0.079&	0.226&	0.182&	0.089&	0.106&	0.117&	0.153&	0.095&	0.093&	0.140&	0.123\\
RS-Net & \textbf{0.061}&	0.166&	0.064&	0.243&	0.170&	\textbf{0.074}&	\textbf{0.063}&	0.098&	0.133&	0.072&	0.103&	0.120&	0.108\\
PointConv &0.082&	0.177&	0.089&	0.237&	\textbf{0.116}&	0.089&	0.094&	0.107&	\textbf{0.124}&	0.061&	0.076&	0.128&	0.110\\
\hline
DGCNN & 0.064&	\textbf{0.160}&	\textbf{0.052}&	\textbf{0.221}&	0.131&	0.085&	0.095&	0.099&	0.127&	\textbf{0.059}&	0.064&	0.118&	\textbf{0.099}\\
GraphCNN & 0.115&	0.178&	0.117&	0.245&	0.160&	0.121&	0.132&	0.115&	0.170&	0.089&	0.098&	0.191&	0.136\\
\hline
Minkowski &0.123&	0.195&	0.100&	0.252&	0.203&	0.140&	0.151&	0.126&	0.154&	0.101&	0.112&	0.154&	0.146\\
\hline
SHOT &0.311&	0.389&	0.193&	0.390&	0.551&	0.350&	0.413&	0.343&	0.276&	0.395&	0.606&	0.374&	0.407\\
\hline
Random &0.329&	0.410&	0.426&	0.452&	0.547&	0.369&	0.488&	0.408&	0.315&	0.396&	0.544&	0.377&	0.446\\
\hline
\end{tabular}}
\end{center}   
\caption{\textbf{Mean Geodesic Error (mGE) results}.}
\label{tab:res1}
\end{table}

In this section, we demonstrate that our proposed method can effectively learn point-wise dense embeddings from human labeled correspondences. We evaluate the embeddings with mGE error. Seven state-of-the-art neural network backbones are benchmarked. These backbones are point cloud~\cite{qi2017pointnet,qi2017pointnet++,pointconv}, graph~\cite{wang2019dynamic,defferrard2016convolutional} and voxel~\cite{choy20194d,tombari2010unique} based neural networks. We additionally compare our approach, which is based on implicit correspondences, with that based on explicit part-level supervision. 



\paragraph{Evaluation and Results}
\label{evaluation}
We split our dataset into train (70\%), validation (15\%) and test (15\%) set. Train and validation sets are used during training and all the results are reported on the test set. We use ADAM optimizer~\cite{kingma2014adam} with initial learning rate $\alpha = 0.001$, $\beta_1 = 0.9$, $\beta_2 = 0.999$ and batch size 4. The learning rate is multiplied by 0.9 every 10 epochs and the hyperparameter $\lambda$ in Equation~\ref{eq:final} is set to 1. The output point embedding vector is 128-dimensional for all neural networks.

Table~\ref{tab:res1} gives mGE of all the compared architectures. SHOT fails to predict correct semantic correspondences across objects, whose performance is just slightly better than random point embeddings. The reason is that SHOT only considers local geometric properties, without aggregation of the global structure and semantic information. 
In contrast, all deep learning based methods using our geodesic consistency loss achieve much smaller mGE. Among them, DGCNN, PointNet, RS-Net and PointConv are relatively superior to the other nets on extracting semantic correspondence information. 
\begin{figure*}[!th]
    \centering
    \includegraphics[width=\linewidth]{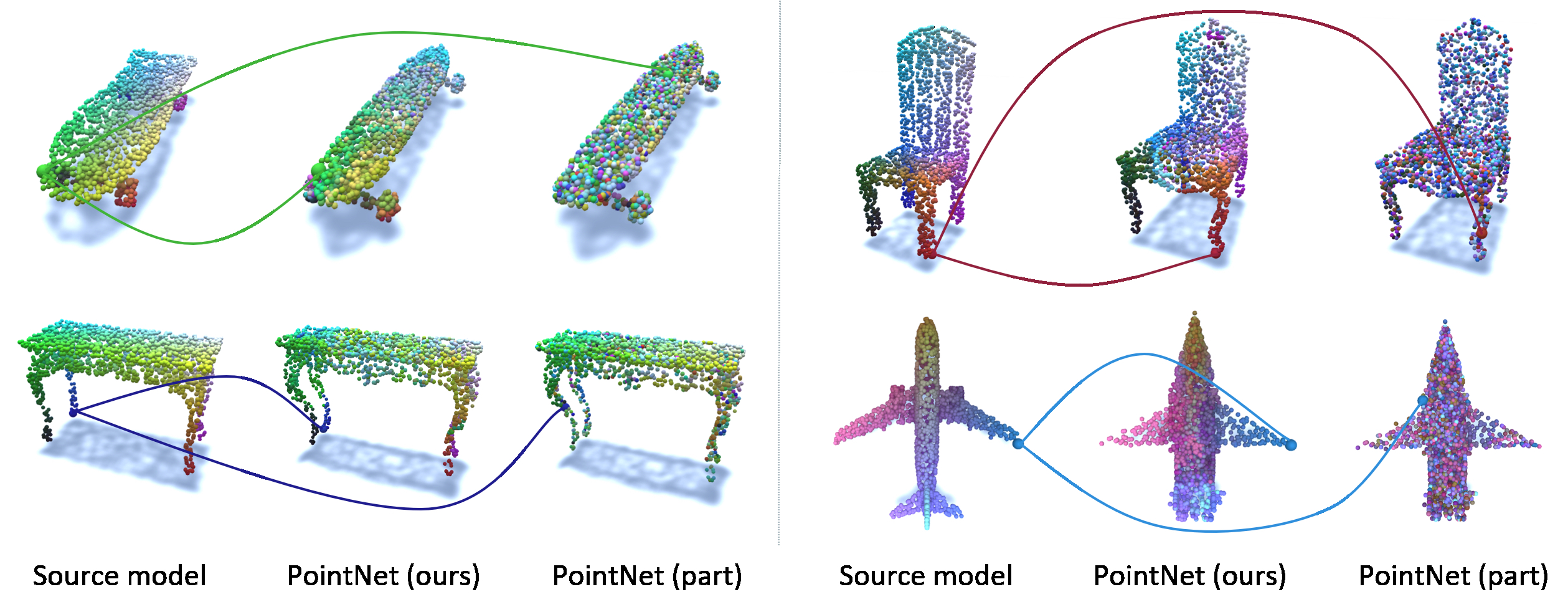}
    \caption{\textbf{Comparison between our method and part-level supervision.} Given a point on the source model, we find its closest point in embedding space on the target model and post-process the founded correspondences with PMF~\cite{vestner2017product} to ensure bijectiveness. The corresponding points are in the same color.}
    \label{fig:part}
\end{figure*}
The visualization of learned embeddings by PointConv is shown in Figure~\ref{fig:emb1}. From Figure~\ref{fig:emb1}, we can see that consistent pointwise embeddings are generated across heterogeneous objects. We get reasonable dense embeddings of all points on objects by fitting sparse correspondence annotations.


\paragraph{Comparison to Part-level Supervision}
To further illustrate the advantage of our proposed semantic correspondence sets, we compare our method with that supervised by part-level annotations. 

We train a PointNet using correspondence labels and part labels respectively. For PointNet trained on part labels, we use the same experiment settings for part segmentation as the original paper~\cite{qi2017pointnet} and extract features from the last but one layer as point embeddings. Then given a point on a source model, we use embeddings to find its corresponding point on the target model and results are shown in Figure~\ref{fig:part}. Qualitatively, we can see that when trained on our correspondence labels, points of the same semantic have similar embeddings while part-level supervision fails to give consistent semantic embeddings across objects. In addition, we compare them quantitatively using mGE, as shown in Table~\ref{tab:part_compare}. Clearly, PointNet trained on our correspondence labels achieves better performance. On the contrary, with only part-level supervision, points in the same part are hard to be distinguished from each other, resulting in inferior performance. Note that the number of training data for part-level supervision (10240) is seven times more than that for correspondence based supervision (1362).

\begin{table}[h]
    \begin{center}
    \resizebox{0.8\textwidth}{!}{
    \begin{tabular}{l|cccccccc}
        \hline
        ~ & Air. & Cap & Car & Chair & Earphone & Guitar & Knife & Lamp \\
        \hline
        PointNet & \textbf{0.063} & \textbf{0.093} & \textbf{0.070} & \textbf{0.088} & 0.151 & \textbf{0.066} & 0.066 & \textbf{0.221} \\
        \hline
        PointNet(Part)  & 0.166 & 0.271 & 0.245 & 0.227 & \textbf{0.140} & 0.083 & \textbf{0.065} & 0.282 \\
        \hline
    \end{tabular}}
    \end{center}
    \begin{center}
    \resizebox{0.8\textwidth}{!}{
    \begin{tabular}{l|ccccccc|c}
        \hline
        ~ & Laptop & Motor & Mug & Pistol & Rocket & Skate. & Table & Average \\
        \hline
        PointNet & 0.163 & \textbf{0.085} & \textbf{0.072} & \textbf{0.091} & \textbf{0.151} & \textbf{0.059} & \textbf{0.042} & \textbf{0.099} \\
        \hline
        PointNet(Part) & \textbf{0.112}  & 0.222 & 0.182 & 0.189 & 0.228 & 0.322 & 0.282 & 0.201 \\
        \hline
    \end{tabular}}
    \end{center}
    \caption{\textbf{Comparison of the results trained on human labeled correspondences and part annotations using PointNet.}}
    \label{tab:part_compare}
\end{table}

\section{Other Applications}
\label{sec:app}
\subsection{Cross-Object Registration}
    We  demonstrate cross-object registration at category-level could benefit from the learnt embeddings, as illustrated in Figure~\ref{fig:registration}. 
    
\noindent\begin{minipage}[b]{0.38\textwidth}
\centering
\includegraphics[width=\textwidth]{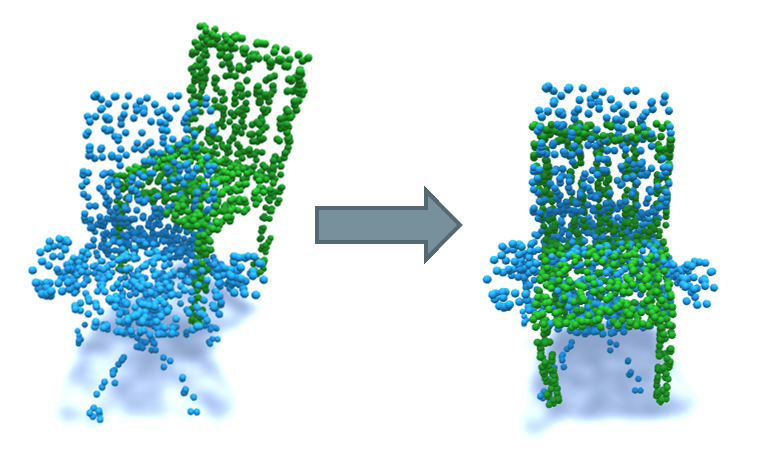}
\captionof{figure}{\textbf{Cross-object registration visualization.}}
\label{fig:registration}
\end{minipage}
\hfill
\begin{minipage}[b]{0.6\textwidth}
\centering
\resizebox{\textwidth}{!}{
\begin{tabular}{l|c|c|c|c}
    \hline
    ~ & Chair & Airplane & Mug & Pistol \\
    \hline
    FPFH   & $\,77.1^\circ/0.285\,$ & $\,41.3^\circ/0.163\,$ & $\,25.9^\circ/0.14\,$ & $\,9.1^\circ/0.095\,$\\
    SHOT   & $\,72.0^\circ/0.262\,$ & $\,44.8^\circ/0.172\,$ & $\,91.3^\circ/0.33\,$ & $\,21.2^\circ/0.121\,$\\
    Part  &  $\,20.1^\circ$\bm{$/0.155\,$} & \bm{$\,24.4^\circ/0.147\,$} & $\,80.6^\circ/0.35\,$ & $\,67.75^\circ/0.306\,$\\
    \hline
    Ours  & \bm{$\,14.6^\circ$}$/0.157\,$ & $\,37.0^\circ/0.225\,$ & \bm{$\,17.1^\circ/0.137\,$} & \bm{$\,5.3^\circ/0.089\,$}\\
    \hline
\end{tabular}}
  \captionof{table}{\textbf{Comparison of cross-object registration.}}
  \label{tab:registration}
\end{minipage}
    
    \paragraph{Experiment Settings} Given two shapes $S$ and $S^\prime$ in the same category with aligned orientations and overlapped centroids, we randomly rotate and translate $S^\prime$. Both shapes are normalized in a unit sphere. The objective is to find a rotation matrix $\mathbf{R} \in \mathbb{R}^{3\times3}$ and a translation vector $\mathbf{t} \in \mathbb{R}^3$ that best align $S$ to $S^\prime$. Initial rotation and translation on $S^\prime$ are seen as ground truth.
    We use RANSAC\cite{fischler1981random} with embeddings for global registration and ICP\cite{besl1992method} to refine. As a comparison, we also evaluate registration results from SHOT, FPFH\cite{rusu2009fast} and PointNet part segmentation embeddings. 840 shape pairs from 4 common categories of CPNet test set are evaluated under three levels of perturbation similar to \cite{pomerleau2015review}: $Easy(10^\circ,0.1)$, $Medium(20^\circ,0.3)$,  $Hard(45^\circ,0.5)$. Table~\ref{tab:registration} gives relative rotational and translational errors. Our embeddings are robust in registration and give reliable semantic correspondences.



\subsection{Partial Object Matching}
In real applications, occlusion and incompletion of 3D models are pretty common, which makes accurate semantic point matching a tough task. 



\begin{figure}[ht]
\begin{center}
  \includegraphics[width=\linewidth]{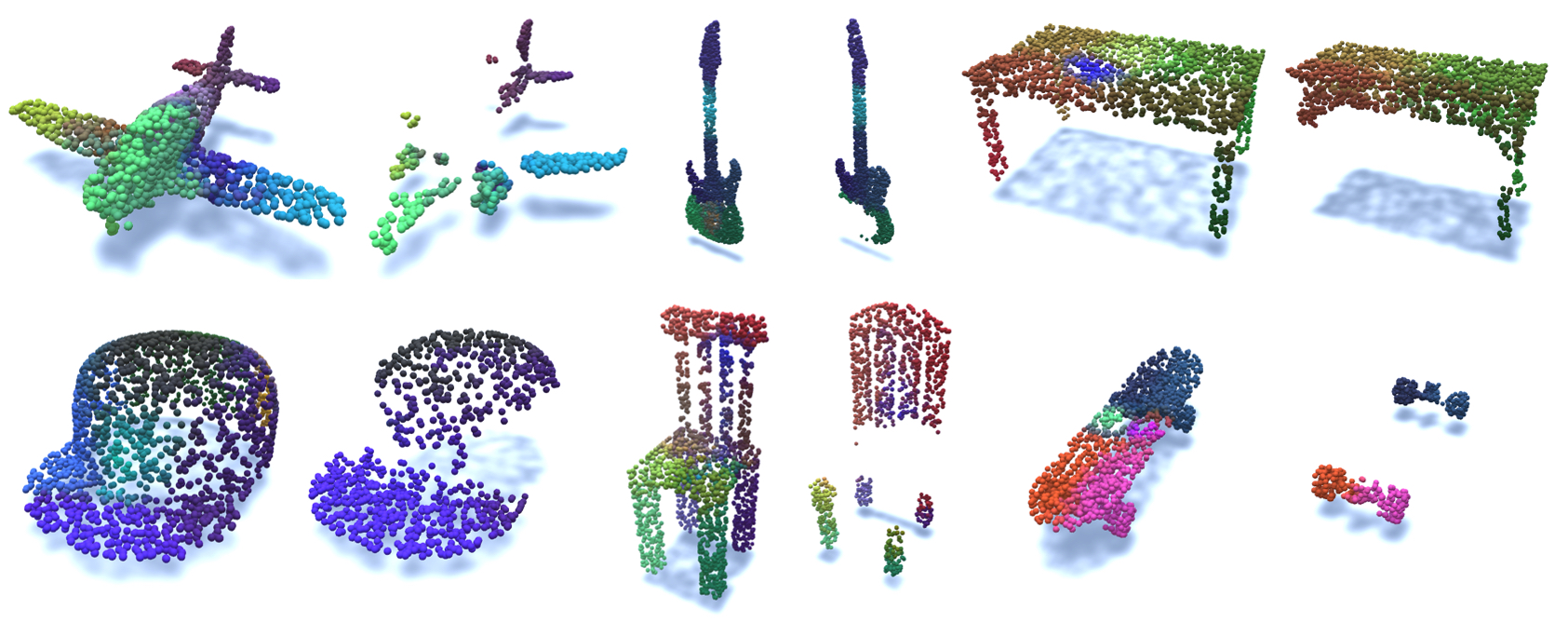}
\end{center}
\caption{\textbf{Partial object matching}. Each pair includes the complete and partial scans of the different objects within the same category. Same colors indicate same embeddings.}
\label{fig:matching}
\end{figure}

We conduct experiments to qualitatively show that the learnt embeddings with our method can generalize well to partial objects and thus can be used to find correspondences between partial and complete objects.  Given our dataset, we train the network with complete objects and apply the network on their partial counterparts synthetically by removing some parts. Figure~\ref{fig:matching} shows the embeddings of partial and complete object pairs. Our method predicts reliable semantic embeddings even under severe erosion. 

\section{Conclusion}

In this paper, we explored a new way towards semantic understanding of 3D objects. Instead of explicitly defining semantic labels on an object, we leveraged an observation that while semantic meanings on a single object can be ambiguous and hard to depict, the correspondences of certain points across objects are clear. We thus built a dataset named \textbf{C}orres\textbf{P}ondence\textbf{Net} (CPNet) based on human labeled correspondences, and proposed a method on learning dense semantic embeddings of objects. Mean Geodesic Error is introduced to evaluate our method with various backbones. Some other applications like cross-object registration and partial object matching are also introduced to better illustrate CPNet's potentiality in boosting general object semantic understandings. 
\section{Acknowledgements}
This work is supported in part by the National Key R\&D Program of China, No. 2017YFA0700800, National Natural Science Foundation of China under Grants 61772332, SHEITC (2018-RGZN-02046) and Shanghai Qi Zhi Institute.
\clearpage

%
%
\bibliographystyle{splncs04}
\bibliography{egbib}
\end{document}